\documentclass[runningheads]{llncs}
\usepackage[T1]{fontenc}
\usepackage{graphicx}
\usepackage{booktabs}
\usepackage{amsmath}
\usepackage{multirow}
\usepackage{array}
\usepackage{pifont}
\usepackage{hyperref}
\newcommand{\cmark}{\ding{51}} 
\usepackage{svg}
\usepackage{float}
\begin{document}
\title{Enhancing Document VQA Models via Retrieval-Augmented Generation}
%
%
\author{
Eric López\inst{1}\orcidID{0009-0003-3229-5739} \and \\
Artemis Llabrés\inst{1}\orcidID{0000-0002-6128-1796} \and \\
Ernest Valveny\inst{1}\orcidID{0000-0002-0368-9697}
}
\authorrunning{E. López et al.}
%
\institute{
Computer Vision Center, Universitat Autònoma de Barcelona, Spain\\
\email{Eric.LopezCe@autonoma.cat, allabres@cvc.uab.cat, ernest@cvc.uab.cat}
}
\maketitle              
\begin{abstract}
Document Visual Question Answering (Document VQA) must cope with documents that span dozens of pages, yet leading systems still concatenate every page or rely on very large vision-language models, both of which are memory-hungry. Retrieval-Augmented Generation (RAG) offers an attractive alternative, first retrieving a concise set of relevant segments before generating answers from this selected evidence. In this paper, we systematically evaluate the impact of incorporating RAG into Document VQA through different retrieval variants—text-based retrieval using OCR tokens and purely visual retrieval without OCR—across multiple models and benchmarks. Evaluated on the multi-page datasets MP-DocVQA, DUDE, and InfographicVQA, the text-centric variant improves the "concatenate-all-pages" baseline by up to +22.5 ANLS, while the visual variant achieves +5.0 ANLS improvement without requiring any text extraction. An ablation confirms that retrieval and reranking components drive most of the gain, whereas the layout-guided chunking strategy—proposed in several recent works to leverage page structure—fails to help on these datasets. Our experiments demonstrate that careful evidence selection consistently boosts accuracy across multiple model sizes and multi-page benchmarks, underscoring its practical value for real-world Document VQA.

\end{abstract}
\noindent\textbf{Project page:} \url{https://pikurrot.github.io/RAG-DocVQA/}
\section{Introduction}
Document VQA aims to enable models to answer questions about the contents of documents, requiring multimodal reasoning over textual, spatial, and visual elements. While early research mainly addressed single-page scenarios through popular benchmarks such as DocVQA, the majority of real-world documents—including manuals, scientific papers, and technical reports—extend across multiple pages. Addressing this gap, MP-DocVQA was recently proposed as a multi-page extension of the original DocVQA dataset, posing natural language questions over sequences of up to twenty pages, accompanied by OCR-extracted text and page images.

Existing state-of-the-art multi-page models, such as Hi-VT5, leverage hierarchical transformer architectures to summarize individual pages before synthesizing the final answer. However, their performance deteriorates as the document length grows, since cross-page reasoning significantly increases complexity. While Large Visual Language Models (LVLMs) such as Qwen2.5-VL can handle long-context inputs, they require substantial computational resources and GPU memory, making deployment costly and inefficient.

A promising alternative is Retrieval-Augmented Generation (RAG), which has recently demonstrated effectiveness in open-domain question answering by first retrieving compact subsets of relevant evidence, thus drastically reducing the inference cost and noise introduced by irrelevant content. In the context of multi-page Document VQA, a RAG pipeline involves segmenting documents into concise textual chunks or visual patches, retrieving the most pertinent segments based on the question, and subsequently generating answers conditioned solely on these retrieved segments.

In this paper, we hypothesize that augmenting existing small and medium-sized Document VQA models with RAG can significantly mitigate the challenges posed by multi-page scenarios. Specifically, we argue that this approach yields performance improvements in both accuracy and computational efficiency across various models and datasets. To validate this hypothesis, our contributions are as follows:
\begin{itemize}
    \item We propose and implement textual RAG variants for two representative Vision Language Models: VT5 and Qwen2.5-VL-7B-Instruct, along with a visual counterpart for Pix2Struct, an OCR-free vision transformer model.
    \item Through comprehensive experiments on MP-DocVQA, DUDE, and InfographicVQA, we demonstrate consistent improvements in question-answering accuracy, ANLS scores, and retrieval metrics, thereby validating our hypothesis.
\end{itemize}

\section{Related Work}
\subsection{Datasets and Task Evolution}
Document Visual Question Answering was initially formalized with the DocVQA dataset \cite{Mathew2021DocVQA}, comprising 50,000 questions posed over more than 12,000 industry-related single-page document images, where answers are explicitly present in the OCR-extracted text. InfographicVQA \cite{Mathew2022InfographicVQA} expanded the task visually, introducing infographic-rich images accompanied by questions that occasionally require numerical or categorical reasoning beyond mere text extraction. Subsequently, DocCVQA \cite{Tito2021DocCVQA} introduced a retrieval-oriented scenario, where questions are answered by retrieving relevant evidence from a collection of 14,362 single-page documents.
Recognizing the limitations of single-page scenarios, MP-DocVQA \cite{Tito2023HiVT5} extended the original DocVQA dataset to multi-page contexts by augmenting each document-question pair with preceding and following pages, forming sequences of up to 20 pages. Concurrently, DUDE \cite{VanLandeghem2023DUDE} provided diverse multi-page documents from various domains, featuring questions that require multi-step reasoning and both extractive and abstractive answers. Additionally, the recent MMLongBench-Doc dataset \cite{Ma2024MMLongBenchDoc} features 1,091 questions over 135 long PDF documents with rich layouts, where 33\% of the questions require cross-page reasoning and 22.5\% are unanswerable, which tests model robustness.

\subsection{From OCR-based to OCR-free Approaches}
Initial Document VQA models largely relied on OCR-extracted text, adapting pre-trained language models such as BERT \cite{Devlin2019BERT} with a classification head to predict the start and end indices where the answer appeared in the text, and T5 \cite{Raffel2020T5} for generative extractive question answering from OCR tokens. LayoutLM \cite{Xu2020LayoutLM} introduced spatial understanding by augmenting textual embeddings from BERT with 2-dimensional positional information derived from OCR bounding boxes. LayoutLMv2 \cite{Xu2021LayoutLMv2} then integrated visual embeddings from the document image and added relative positional biases to the self-attention mechanism. LayoutLMv3 \cite{Huang2022LayoutLMv3} further refined this by encoding images as patch-level embeddings.
To further leverage visual modalities, DocFormer \cite{Appalaraju2021DocFormer} combined textual, visual, and spatial features via a novel multi-modal self-attention mechanism, using a ResNet \cite{He2016ResNet} backbone for image features. Later, DocFormerv2 \cite{Appalaraju2023DocFormerv2} replaced the ResNet CNN with a linear projection of image patches and adopted a T5-based encoder-decoder architecture.
Addressing multi-page reasoning, Hi-VT5 \cite{Tito2023HiVT5} introduced a hierarchical transformer where the encoder summarizes each page independently, and the decoder attends to these summaries to generate the answer. GRAM proposed inserting additional document-level transformer layers between page-level encodings, using document tokens to propagate information across pages, achieving state-of-the-art performance when combined with DocFormerv2.
More recently, OCR-free models emerged, eliminating explicit text extraction. Pix2Struct \cite{Lee2023Pix2Struct} proposed a fully end-to-end approach without explicit OCR, pretraining a vision transformer encoder to transcribe masked web screenshots into simplified HTML. Nevertheless, handling multi-page scenarios remained challenging due to growing computational complexity.

\subsection{Generic Vision-Language Models}
Recently, generic vision-language models (VLMs), such as Qwen2.5-VL \cite{Bai2025Qwen2.5VL}, have started being explored for Document VQA tasks. These models can directly process combined visual and textual inputs thanks to extensive pre-training on large multimodal datasets. However, a major issue with these models is their high computational cost and GPU memory usage, making practical deployment difficult. To mitigate such limitations, retrieval-based strategies like ColPali \cite{Faysse2025ColPali} have been adopted, which extends the PaliGemma \cite{Beyer2024PaliGemma} vision-language model with a ColBERT-style \cite{Khattab2020ColBERT} late interaction mechanism, generating visual page embeddings optimized for retrieval tasks, thus improving evidence selection prior to answer generation.

\section{Methodology}
\label{methodology}

We adapt Retrieval-Augmented Generation (RAG) to multi-page Document VQA through a three-stage pipeline comprising indexing, retrieval, and generation. In particular, we propose two complementary variants: a textual RAG, applied to two language-based models—a smaller encoder–decoder (VT5) and a larger vision–language model (Qwen2.5-VL-7B)—and a visual RAG, where retrieval and generation operate directly on image inputs without explicit OCR, using Pix2Struct as our baseline model.

\subsection{Textual RAG}

\begin{figure}
\includegraphics[width=\textwidth]{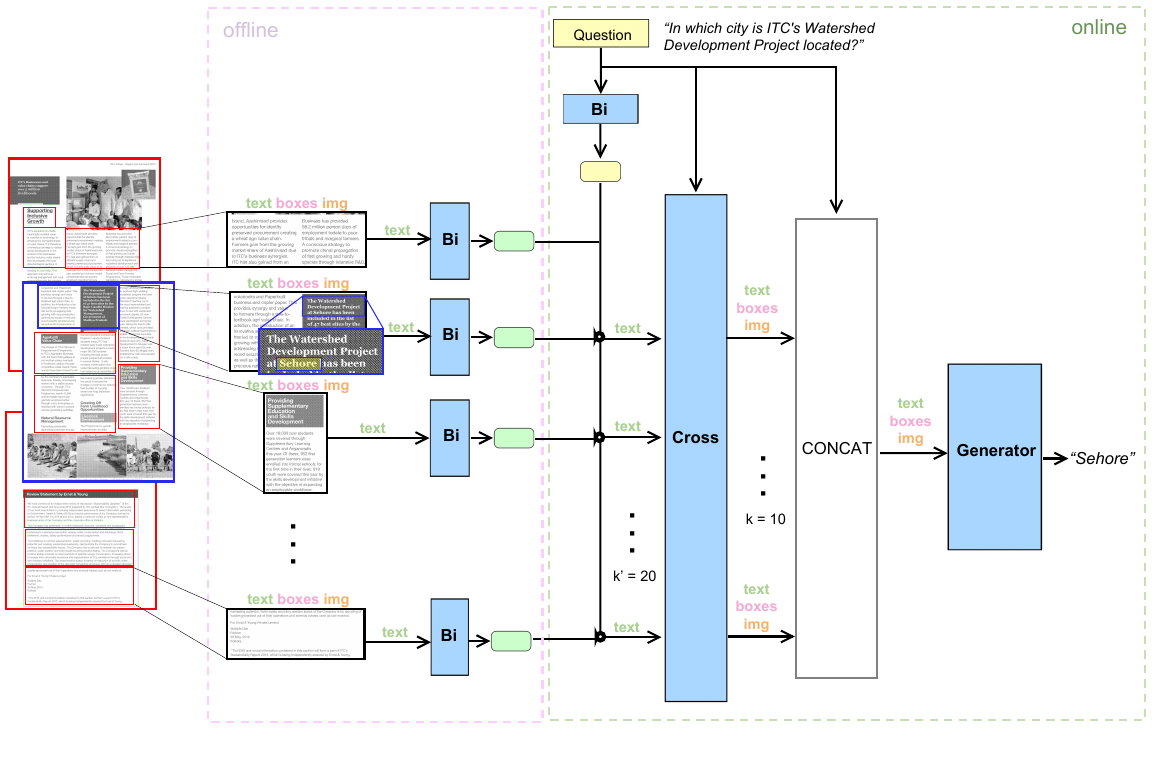}
\caption{Overview of the \textbf{Textual RAG} pipeline. \textbf{Offline (pink panel):} A multi-page document is segmented into chunks consisting of OCR (text and boxes) and the image crop of the chunk. A Bi-Encoder converts the chunk text into a dense embedding for later retrieval. \textbf{Online (green panel):} the user question is encoded in the same way, and cosine-similarity is used to select the top-$k'$ chunks, whose text is then passed to a Cross-Encoder (Reranker) for a more refined filtering and ranking. The full information of the highest-ranked $k$ chunks is then concatenated and fed to the Generator to produce an answer.} \label{fig:text_rag}
\end{figure}

\subsubsection{Indexing}
We first segment the OCR-extracted token sequences of the multi-page documents into textual chunks. The chunking process is controlled by three parameters: chunk size $L$ (tokens per chunk), overlap $O$ (shared tokens between neighbors), and chunk-size tolerance $\tau$, which allows the last chunk on a page to expand to \((1 + \tau) \times L\) tokens, so that small remainders are merged rather than isolated.
Each chunk text is then encoded offline into a fixed-dimensional vector embedding using a bi-encoder architecture. Specifically, we use \textit{bge-en-small-v1.5} \cite{BAAI2023bge-small}, which has proven to be the most effective and efficient. We fine-tune this embedding model using contrastive learning on dataset-specific query–chunk pairs, optimizing a Multiple-Negatives Ranking Loss (details in Section \ref{experimental_setup}). Finally, we store each chunk's tokens, bounding boxes, image crop (extracted from the chunk bounding box), and embedding for fast retrieval.
\subsubsection{Retrieval}
At inference time, we encode the input query ($q$) using the same embedding model with shared weights, obtaining an embedding vector of the same dimensionality as the chunks. We compute the cosine similarity between the query embedding and all stored chunk embeddings, retrieving an initial candidate set of the top $k'$ most similar chunks. To further enhance precision, we employ an additional cross-encoder reranker—a transformer-based classification model trained specifically to discriminate between relevant and irrelevant query–chunk pairs. Unlike the initial bi-encoder, the cross-encoder reranker jointly encodes each query together with each candidate chunk in a single forward pass, leveraging full self-attention across all tokens from both inputs. It outputs a refined relevance score indicating the likelihood that the chunk contains the answer. Finally, we select the top $k$ chunks with the highest reranked scores, which are then used for generation.
\subsubsection{Generation}
The generation step differs according to the underlying model used. For VT5, the $k$ top-ranked chunks' text tokens are concatenated with the question tokens to form the semantic input. Question tokens are assigned placeholder bounding boxes with all coordinates set to zero and concatenated with the chunks’ original boxes, guaranteeing that the semantic and spatial sequences remain aligned. The image crops of the different chunks are arranged into a compact adaptive grid, minimizing the total area and thereby optimizing input efficiency. As in Hi-VT5, the three input modalities—semantic, spatial, and visual—are first embedded separately. The semantic and spatial information of each token is fused by summing their corresponding embeddings:

\begin{equation}
\mathcal{E}_i = E_O(O_i) + E_x(x^i_0) + E_y(y^i_0) + E_x(x^i_1) + E_y(y^i_1)
\label{eq:eps_fusion}
\end{equation}

where  
\(E_O(O_i)\) is the semantic embedding of OCR token \(O_i\), produced by a T5 language backbone, and  
\(E_x\), \(E_y\) are learned embeddings for the bounding box coordinates.

\begin{figure}[H]
\includegraphics[width=\textwidth]{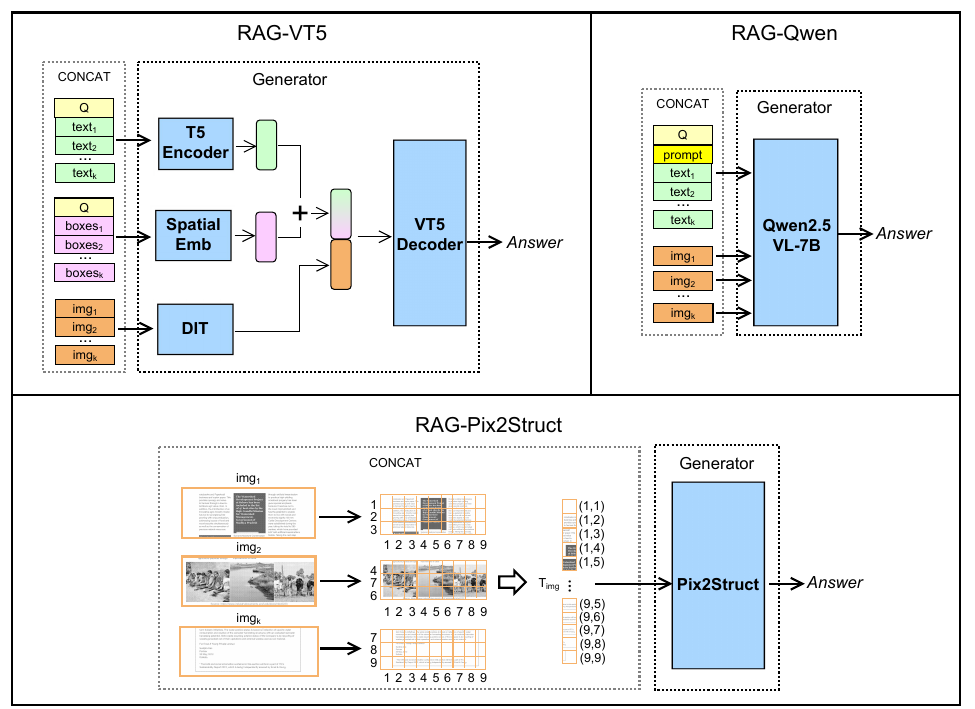}
\caption{Different concatenation and generation strategies for the proposed RAG methods. For RAG-VT5, the information of different modalities is separately concatenated and fed to a specialized encoder, before aggregating and passing it to the decoder. For RAG-Qwen, the text of the chunks is concatenated along with the question and an instruction prompt, and is passed together with the separate chunk image crops to the model. For RAG-Pix2Struct, only the image crops are processed, by tiling them into mini-patches, assigning a positional coordinate to each one and feeding them to the model.} \label{fig:generators}
\end{figure}

The image crops grid is encoded into an embedding $V$ by a Document Image Transformer (DIT), specifically \textit{dit-base-finetuned-rvlcdip}.
We then concatenate all the question embeddings \(\mathcal{E}^q\), the text embeddings \(\mathcal{E}^{o}\), and the visual patch embeddings \(V\) to form the final input sequence \([\mathcal{E}^q;\, \mathcal{E}^{o};\, V]\) which is fed to the VT5 decoder to generate the answer. This generation process for VT5 is illustrated in Fig.~\ref{fig:generators} (top-left).

For Qwen2.5-VL-7B, the retrieved chunks' texts are concatenated with the question and a prompt (\textit{"Directly provide only a short answer to the question."}) to form a single textual input sequence. Each chunk's corresponding image crop is embedded independently by Qwen’s vision encoder and integrated within the textual input through special placeholders. The combined multimodal embeddings are then processed by Qwen’s decoder, autoregressively generating the final answer. The generation procedure for Qwen2.5-VL-7B is depicted in Fig.~\ref{fig:generators} (top-right).

\subsection{Visual RAG}

\begin{figure}
\includegraphics[width=\textwidth]{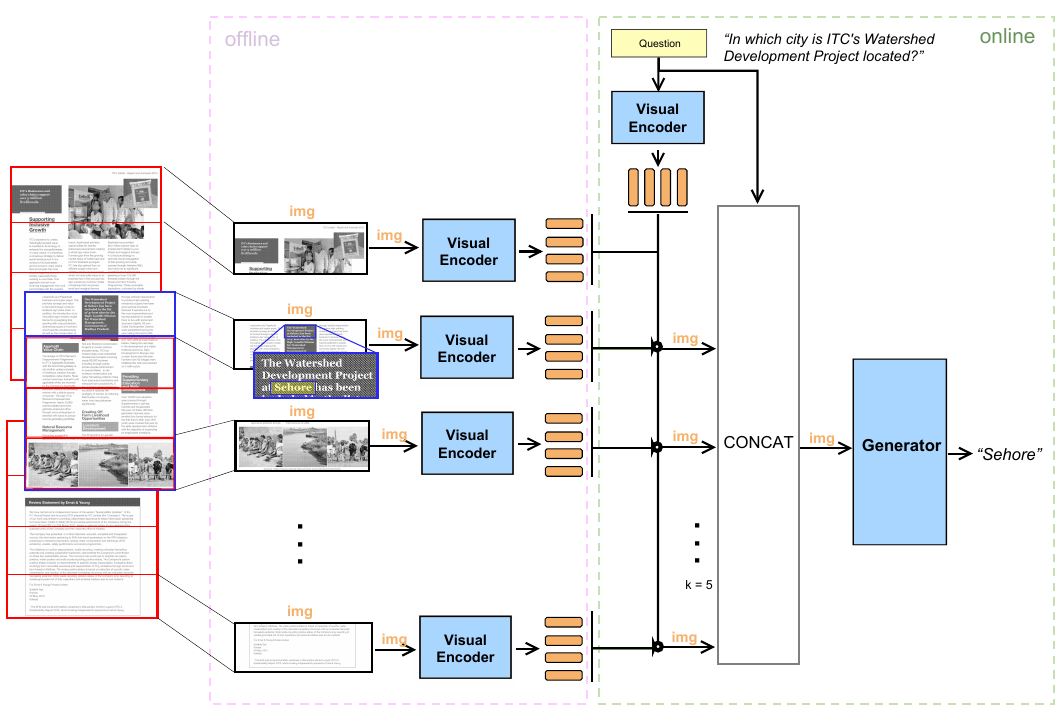}
\caption{Overview of the \textbf{Visual RAG} pipeline. \textbf{Offline (pink panel):} A multi-page document is segmented into image patches, each of which is passed to a Visual Encoder to produce multi-vector embeddings for later retrieval. \textbf{Online (green panel):} the user question is rendered as an image, encoded in the exact same way and matched to the patch embeddings through late interaction. The top-$k$ image patches are concatenated (see Fig.~\ref{fig:generators} (bottom)) and passed to the generator.} \label{fig:visual_rag}
\end{figure}

\subsubsection{Indexing}
Document images are horizontally segmented into overlapping visual patches. Each patch has a fixed vertical pixel size (patch size $P$), and overlaps by half of this height with adjacent patches to ensure no crucial information is lost across patch boundaries. Each visual patch is encoded independently into multi-vector embeddings $E_p$ of shape \(T_{\text{img}} \times 768\) using a Visual Encoder, where $T_{\text{img}}$ is the number of image tokens. Specifically, we use the Pix2Struct vision transformer encoder, which works with an image sequence length of \(T_{\text{img}}=2048\).
\subsubsection{Retrieval}
For retrieval, the textual query is rendered as an image on a white background and embedded into multi-vector representations $E_q$ of shape \(T_{\text{img}} \times 768\) using the same Pix2Struct encoder as used for the document patches. We perform late interaction retrieval following ColBERT~\cite{Khattab2020ColBERT}, which computes token-level similarity $S_{q, p}$ between the query $q$ and each candidate patch $p$, allowing fine-grained matching:

\begin{equation}
S_{q, p} = \sum_{i \in [|E_q|]} \max_{j \in [|E_p|]} E_{q_i} \cdot E_{p_j}
\end{equation}

The top-$k$ most relevant patches are selected based on their similarity scores. If retrieved patches overlap spatially, overlapping regions are merged into single unified patches, removing duplicate visual information.

\subsubsection{Generation}
The concatenation and generation process for RAG-Pix2Struct is shown in Fig.~\ref{fig:generators} (bottom). The retrieved $k$ image patches are resized and concatenated to form the input sequence for Pix2Struct \cite{Lee2023Pix2Struct}. Specifically, each image patch is scaled so it can be evenly divided into small, non-overlapping $16\times16$ mini-patches. These mini-patches receive 2-dimensional positional indices: row indices increase continuously across the concatenated patches (from top to bottom), while column indices restart at 1 within each patch (from left to right). This indexing preserves each patch's internal 2-D layout while forming a vertical stack of patches. Finally, the resulting sequence of \(T_{\text{img}}\) mini-patches is fed directly into Pix2Struct, which encodes them using its built-in 2-D positional encoder to generate the final answer.

\section{Experimental Setup}
\label{experimental_setup}
\subsection{Evaluation Metrics}
We employ metrics for two complementary aspects of the task: (i) question answering, and (ii) retrieval quality of relevant document content.

\paragraph{(i) Question Answering Metrics.}
To evaluate the final output of the generator, we use standard accuracy and ANLS (Average Normalized Levenshtein Similarity), which are commonly used in Document VQA benchmarks to measure the correctness and textual similarity of predicted answers with respect to ground truth.

\paragraph{(ii) Retrieval Quality Metrics.}
To assess the effectiveness of our retrieval pipeline, we use two additional metrics:

\textit{• Page-level retrieval.}
We adapt the retrieval precision metric, originally used by existing multi-page models to evaluate their page-retrieval modules, to measure whether any of the retrieved chunk comes from the correct page. Specifically, we define Retrieval Precision@$k$ as:
\begin{equation}
\text{Retrieval Precision}@k \;=\;
\mathbf{1}\!\left[\;
\exists\, j \in \{1,\dots,k\}\;:\; p_{\text{gt}} = p^{\text{pred}}_j
\right]
\end{equation}
where \(p_{\text{gt}}\) is the ground-truth answer page and \(p^{\text{pred}}_{j}\) is the page associated with the \(j\)-th retrieved chunk (\(j=1,\dots,k\)).

\textit{• Chunk-level semantic matching.}
We introduce Chunk Score@$k$ as a smooth measure of how similar the ground-truth answer is to any part of the retrieved text, even if not an exact match:
\begin{equation}
\text{Chunk Score}@k = \log_2\left(1 + \max_{1 \leq j \leq k} \,\text{sim}(c_j, a)\right)
\end{equation}
where \(\text{sim}(c_j, a)\) computes the similarity between chunk \(c_j\) and the ground-truth answer \(a\). This similarity is defined as the log-scaled maximum inverse-edit-distance between \(a\) and all possible substrings of equal length within \(c_j\).

\subsection{Implementation Details}

The specific hyperparameter values shown in Table~\ref{tab:hparams} are the ones that yield better results in our experimentation. For the bi-encoder, we used bge-en-small-v1.5 \cite{BAAI2023bge-small} to encode both the query and the document chunks, and for the cross-encoder we used bge-reranker-v2-m3 \cite{BAAI2023bge-reranker1,BAAI2023bge-reranker2} to rerank and filter the retrieved chunks.

To enable a fair multi-page comparison between our RAG variants and their single-page baselines, we extend each baseline with a lightweight multi-page strategy. For VT5, we adopt a “concatenate” approach analogous to the T5-concat method in \cite{Tito2023HiVT5}, merging the OCR tokens—and, in our case, the page screenshots—of all pages into a single input context. For Pix2Struct, concatenating complete page images would over-shrink text, so we instead process each page independently, generate an answer per page, and retain the answer with the highest confidence; we refer to this variant as Pix2Struct-MaxConf.

\begin{table}[t]
  \caption{Key hyper-parameters used for the Textual RAG models (left) and the visual RAG (right).}
  \label{tab:hparams}
  \centering

  \renewcommand{\arraystretch}{1.3}
  \begin{minipage}[t]{0.45\linewidth}
    \centering
    \textbf{RAG-VT5 / RAG-Qwen}\par\vspace{4pt}
    \setlength{\tabcolsep}{8pt}
    \begin{tabular}{|l|l|}
      \hline
      Parameter & Value \\
      \hline
      Chunk size $L$ & 60 \\
      \hline
      Chunk size Tolerance $\tau$ & 0.2 \\
      \hline
      Overlap $O$ & 10 \\
      \hline
      $k'$ & 20 \\
      \hline
      $k$ & 10 \\
      \hline
    \end{tabular}
  \end{minipage}
  \hfill
  \begin{minipage}[t]{0.45\linewidth}
    \centering
    \textbf{RAG-Pix2Struct}\par\vspace{4pt}
    \setlength{\tabcolsep}{8pt}
    \begin{tabular}{|l|l|}
      \hline
      Parameter & Value \\
      \hline
      Patch size $P$ & 512 \\
      \hline
      Overlap $O_{pix}$ & 256 \\
      \hline
      $k$ & 5 \\
      \hline
    \end{tabular}
  \end{minipage}
\end{table}

\subsection{Training}
The bi-encoder embedding model is fine-tuned using contrastive learning. For this, we create a synthetic dataset of anchor-positive pairs, where each anchor is a query and the corresponding positive is a chunk that previously yielded good retrieval performance. Specifically, we first process each document VQA dataset through our RAG pipeline with the original embedding model. For each retrieved set of $k$ chunks, we individually pass each chunk to the generator, obtaining 
$k$ separate answers. These answers are compared with the ground truth, and the chunk producing the highest ANLS is selected for that sample. Query–chunk pairs that achieve an ANLS greater than a threshold $t$, which we empirically set to $0.8$, form the positive training set. The bi-encoder is subsequently fine-tuned using a Multiple Negatives Ranking Loss \cite{Henderson2017SmartReply}, where for a given positive pair (query $q_1$, chunk $c_1$), the other chunks $c_2, ..., c_b$ in the batch serve as negatives.

Due to the sparse, discontinuous, and contextually diverse nature of retrieval-augmented inputs, we further fine-tune the generator models on each target VQA dataset, keeping the bi-encoder embedding weights frozen. For VT5, we perform full fine-tuning of all layers—including the language backbone and the spatial and visual embedding modules—using the AdamW optimizer with a linear learning-rate scheduler, an initial learning rate of $2e-4$, 1000 warm-up steps, and 4 epochs. For Qwen2.5-VL-7B-Instruct, we employ parameter-efficient Low-Rank Adaptation (LoRA), applying it specifically to the query and value projection matrices with parameters $\alpha=16$, $\text{rank}=8$ and $\text{dropout}=0.05$. Training for both the embedding model and VT5 was performed on an NVIDIA TITAN Xp GPU (12 GB). The Qwen LoRA fine-tuning was conducted on an NVIDIA L40S GPU (46 GB). For Pix2Struct, we use versions already fine-tuned on each target dataset, so no further training was required.

\section{Results}

\subsection{Quantitative Results}
\begin{table}[t]
\setlength{\tabcolsep}{5pt}
\centering
\caption{Comparison of published multi-page Document VQA systems, our multi-page
baselines, and their retrieval-augmented variants.\label{tab:results_sota}}
\vspace{4pt}
\begin{tabular}{lccccc}
\toprule
\textbf{Method} & \textbf{Params (M)} &
\multicolumn{3}{c}{\textbf{ANLS (\%)}}\\
\cmidrule(lr){3-5}
 & & MP-DocVQA & DUDE & InfoVQA\\
\midrule
\multicolumn{5}{l}{\hspace{1em}\textit{Published SOTA}}\\
T5-concat\cite{Tito2023HiVT5}                           & 223     & 50.5 & 38.7 & --- \\
Pix2Struct-SA-retrieval\cite{Kang2024SelfAttnMPDocVQA}             & 273     & 62.0 & ---  & --- \\
Hi-VT5\cite{Tito2023HiVT5}                              & 316     & 61.8 & 35.7 & --- \\
GRAM\cite{Blau2024GRAM}                                & 281     & \textbf{73.9} & \textbf{46.2} & --- \\
\midrule
\multicolumn{5}{l}{\hspace{1em}\textit{Multi-page baselines}}\\
Pix2Struct-baseline                        & 282     & 49.1 & 18.9 & 30.2 \\
VT5-baseline                               & 223     & 50.0 & 36.0 & 21.1 \\
Qwen2.5-VL-baseline                        & 7 000   & 51.2 & 35.9 & 61.6 \\
\midrule
\multicolumn{5}{l}{\hspace{1em}\textit{Retrieval-augmented variants}}\\
RAG-Pix2Struct (ours)                      & 282            & 54.1 & 18.1 & 33.5 \\
RAG-VT5 (ours)\textsuperscript{\dag}       & 223 + 601      & 63.1 & 43.3 & 31.7 \\
RAG-Qwen2.5-VL (ours)\textsuperscript{\dag}& 7 000 + 601 & 73.7 & 39.9 & \textbf{63.6} \\
\bottomrule
\end{tabular}

\vspace{2pt}
{\footnotesize%
\raggedright%
\textsuperscript{\dag}\;Textual RAG totals include an additional 601M parameters
(33M bi-encoder + 568M cross-encoder) used in the retrieval stack.\par}
\end{table}

Table \ref{tab:results_sota} compares four publicly available models of similar parameter scale, the single-page baselines adapted to multi-page input (Pix2Struct-baseline, VT5-baseline, Qwen2.5-VL-baseline), and their corresponding RAG versions. RAG consistently boosts the textual models: RAG-VT5 gains +13.2, +7.3, and +10.6 ANLS on MP-DocVQA, DUDE, and InfographicVQA, respectively, while RAG-Qwen improves by +22.5, +4.0, and +2.0 on the same datasets. For the visual model, RAG-Pix2Struct yields moderate gains on MP-DocVQA (+5.0 ANLS) and InfographicVQA (+3.3 ANLS) and a slight drop on DUDE (-0.8 ANLS). We attribute this to DUDE’s higher reasoning complexity—scenarios where processing each full-page image (as in the baseline) can retain crucial global context that retrieval-based pruning may omit. Apart from this outlier, the gains are consistent across models and datasets, confirming the general effectiveness of retrieval-augmented generation for multi-page Document VQA.
Unlike Pix2Struct-SA-retrieval, which recomputes full self-attention between the question and page features at inference time, our RAG-Pix2Struct encodes all patches offline and uses a lightweight late-interaction step. While the performance is not as high, this cuts GPU cost and latency, making real-world application more feasible.

\subsection{Qualitative Results}

\begin{figure}[H]
\includegraphics[width=\textwidth]{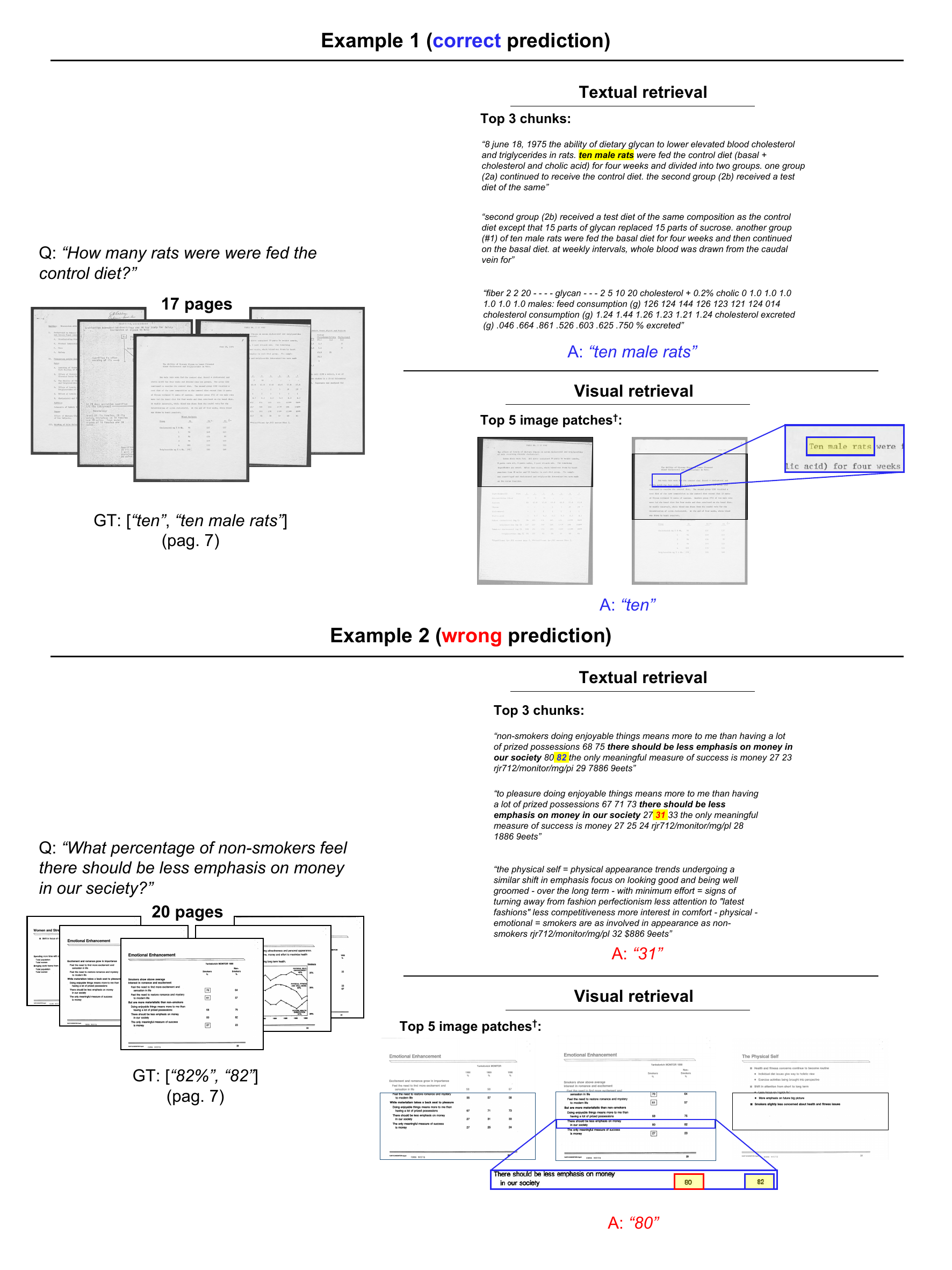}
\caption{Qualitative examples of Question Answering on long-context multi-page samples of MP-DocVQA. Textual RAG uses VT5 as generator, while Visual RAG uses Pix2Struct. Example 1 shows a correct retrieval and correct model generation, while Example 2 shows a correct retrieval but a wrong generation.} \label{fig:qualitative_examples}
{\footnotesize%
\raggedright%
\textsuperscript{\dag}\;While 5 image patches are retrieved, some may be overlapping so they are merged into a bigger one.\par}
\end{figure}

Fig.~\ref{fig:qualitative_examples} illustrates qualitative examples comparing the performance of Textual and Visual RAG on two representative samples. In Example 1, we present a scenario in which a question is posed to a 17-page document. Textual retrieval selects the top relevant text chunks, concatenates them, and passes them to the generator. Similarly, visual retrieval identifies the top image patches. Here, both models generate a correct answer, as the retrieved evidence clearly contains the necessary information.

In contrast, Example 2 depicts a case where, despite accurate retrieval, the generated answers are incorrect. For textual retrieval, the error arises because the retrieved chunks provide ambiguous context; specifically, two different retrieved tables could individually answer the question, but only one is appropriate (the first retrieved chunk corresponds to non-smokers). For visual retrieval, the model fails to identify the correct column within the retrieved image patches, as crucial contextual information—the column headers distinguishing between smoker categories—is missing.

\section{Ablation Study}
\begin{table}[h]
\centering
\caption{Ablation of \textbf{RAG-VT5 on MP-DocVQA validation set}: impact of adding reranking, embedder fine-tuning, layout segmentation, spatial clustering, and layout loss.}
\label{tab:experiments}
\begin{tabular}{
  >{\centering\arraybackslash}p{1.2cm}
  >{\centering\arraybackslash}p{1.2cm}
  >{\centering\arraybackslash}p{1.2cm}
  >{\centering\arraybackslash}p{1.2cm}
  >{\centering\arraybackslash}p{1.2cm} |
  >{\centering\arraybackslash}p{1cm} 
  >{\centering\arraybackslash}p{1cm} 
  >{\centering\arraybackslash}p{1.5cm} 
  >{\centering\arraybackslash}p{1.5cm}
}
\hline
Rerank & Train Emb. & Layout & Cluster & Train Layout & ANLS           & Acc            & Ret. Prec.@$k$  & Chunk Score@$k$\\ \hline
       &            &        &         &              & 58.23          & 50.74          & \textbf{97.20}  & \textbf{98.18} \\
\cmark &            &        &         &              & 61.06          & 53.57          & 95.39           & 97.78          \\
\cmark & \cmark     &        &         &              & \textbf{61.46} & \textbf{53.82} & 96.20           & 97.97          \\
\cmark & \cmark     & \cmark &         &              & 57.92          & 49.01          & 95.41           & 95.55          \\
\cmark & \cmark     & \cmark & \cmark  &              & 58.45          & 49.54          & 96.26           & 95.37          \\
\cmark & \cmark     & \cmark & \cmark  & \cmark       & 59.14          & 49.99          & 96.26           & 95.37          \\ \hline
\end{tabular}
\end{table}

We conduct an ablation study to assess the contribution of individual components in our RAG pipeline, using the RAG-VT5 variant on the MP-DocVQA validation set. Alongside standard performance metrics (ANLS and accuracy), we also report retrieval-specific metrics—retrieval precision and chunk score—as defined in Section~\ref{experimental_setup}. The analysis is divided into two parts: (i) the impact of retrieval enhancements (reranking and embedding fine-tuning), and (ii) the effect of layout-guided chunking.

\subsection{Retrieval variants}
The base retrieval system consists of chunking the document, encoding each chunk with a frozen embedding model, and selecting the top-$k'$ chunks via cosine similarity. To this, we add two enhancements: a cross-encoder reranker that refines the top-$k'$ selection into a more accurate top-$k$ subset, and contrastive fine-tuning of the embedding model to improve the quality of initial retrieval.
As shown in Table~\ref{tab:experiments}, the base pipeline yields the highest retrieval precision, likely because it selects a larger candidate set ($k$), increasing the chance of including the correct chunk. However, this broader context introduces more irrelevant content, which can hurt answer accuracy. Adding the reranker improves ANLS and accuracy by narrowing the context to the most relevant chunks. Further fine-tuning the embedding model does not significantly raise ANLS but leads to improved retrieval precision and chunk score, indicating more effective early filtering.

\subsection{Layout segmentation}
While our default chunks are formed by sliding windows over the raw OCR tokens, several recent works show that layout-aware chunking—in which segments follow the page’s structural regions—can improve retrieval accuracy or downstream document-understanding tasks such as question answering and information extraction \cite{Verma2025S2Chunking,Zhao2024DocLayoutYOLO,Wang2023DocLLM}.

We therefore replace the sliding-window chunks with regions produced by a Document Image Transformer (DIT). Specifically, we use \textit{cmarkea/dit-base-layout-detection} \cite{DeDitLay}, a DIT that outputs 11 fine-grained labels per page. Small boxes or boxes overlapping more than 50\% with larger ones are considered noise and discarded. Labels are simplified to four categories: \textit{title}, \textit{text}, \textit{figure}, and \textit{table}. Centroids of these boxes form an undirected graph with edge weights as inverse Euclidean distances. Spectral clustering is then applied, selecting the number of clusters per page by maximizing the silhouette coefficient. Clusters are merged into rectangular layout chunks.

During training, each OCR token receives a layout embedding scaled by a factor of 10 to match the semantic and spatial embeddings and summed with them. Additionally, a linear head attached to every encoder output predicts each token’s layout label, and the resulting cross-entropy loss is added to the main VQA loss.

Table~\ref{tab:experiments} shows that layout-based chunks hurt retrieval precision and, despite a modest recovery after clustering and the auxiliary loss, never surpass the simpler window strategy. We attribute this to over-segmentation of pages, which fragments the context and prevents the generator to answer with the enough information.

\section{Conclusion}
In this work, we revisited multi-page Document VQA through the lens of Retrieval-Augmented Generation. We adapted RAG pipelines—text-based for VT5 and Qwen2.5-VL, and purely visual for Pix2Struct—to efficiently select and aggregate evidence from long documents. Across three multi-page benchmarks, our approach consistently improved performance over strong baselines, achieving up to +22 ANLS gains while avoiding the memory demands of processing full document sequences. Through ablation studies, we found that reranking and contrastive embedder fine-tuning are key to these gains, whereas layout-guided chunking—despite success in other document tasks—did not yield benefits on our benchmarks. These results demonstrate that careful evidence selection enables smaller models to effectively tackle real-world, multi-page document VQA.

\section*{Acknowledgments}
This research has been supported by the Consolidated Research Group 2021 SGR 01559 from the Research and University Department of the Catalan Government, and by project PID2023-146426NB-100 funded by MCIU/AEI/10.13039/ 501100011033 and FEDER, UE.

This work has also been funded by the European Lighthouse on Safe and Secure AI (ELSA) from the European Union’s Horizon Europe programme under grant agreement No 101070617.

With the support of the FI SDUR predoctoral grant program from the Department of Research and Universities of the Generalitat de Catalunya and co-financing by the European Social Fund Plus (2024FISDU\_00095).

\bibliographystyle{splncs04}
\bibliography{refs}

\end{document}